\documentclass[runningheads]{llncs}
\usepackage{graphicx}
\usepackage{comment}
\usepackage{amsmath,amssymb} 
\usepackage{color}
\usepackage{cite}

\newcommand{\name}{Sat2Graph}

\newcommand{\songtao}[1]{\textcolor{blue}{Songtao:#1}}
\newcommand{\srm}[1]{\textcolor{red}{Sam:#1}}

\newcommand{\san}[1]{\textcolor{green}{Sanjay:#1}}

\newcommand{\dontshow}[1]{}

\begin{document}
\pagestyle{headings}
\mainmatter
\def\ECCVSubNumber{4510}  

\title{\name: Road Graph Extraction \\ through Graph-Tensor Encoding} 

\titlerunning{Sat2Graph}
%

\author{Songtao He\inst{1} \and
Favyen Bastani\inst{1} \and
Satvat Jagwani\inst{1} \and 
Mohammad Alizadeh\inst{1} \and
Hari Balakrishnan\inst{1} \and
Sanjay Chawla\inst{2} \and
Mohamed M. Elshrif\inst{2} \and
Samuel Madden\inst{1} \and
Mohammad Amin Sadeghi\inst{3} 
}
\authorrunning{S. He et al.}
%
\institute{Massachusetts Institute of Technology \\
\email{\{songtao, favyen, satvat, alizadeh, hari, madden\}@csail.mit.edu} \and
Qatar Computing Research Institute \\ \email{schawla@hbku.edu.qa, melshrif77@gmail.com} \and 
University of Tehran \\
\email{m.a.sadeghi@gmail.com}\\
}
\maketitle

\begin{abstract}
Inferring road graphs from satellite imagery is a challenging computer vision task. Prior solutions fall into two categories: (1) pixel-wise segmentation-based approaches, which predict whether each pixel is on a road, and (2) graph-based approaches, which predict the road graph iteratively. We find that these two approaches have complementary strengths while suffering from their own inherent limitations.\\ 

In this paper, we propose a new method, \name, which combines the advantages of the two prior categories into a unified framework. The key idea in \name\ is a novel encoding scheme, {\em graph-tensor encoding} (GTE), which encodes the road graph into a tensor representation. GTE makes it possible to train a simple, non-recurrent, supervised model to predict a rich set of features that capture the graph structure directly from an image. We evaluate \name\ using two large datasets. We find that \name\ surpasses prior methods on two widely used metrics, TOPO and APLS. Furthermore, whereas prior work only infers planar road graphs, our approach is capable of inferring stacked roads (e.g., overpasses), and does so robustly. 

\end{abstract}

\section{Introduction}

Accurate and up-to-date road maps are critical in many applications, from navigation to self-driving vehicles. However, creating and maintaining digital maps is expensive and involves tedious manual labor. In response, automated solutions have been proposed to automatically infer road maps from different sources of data, including GPS tracks, aerial imagery, and satellite imagery. In this paper, we focus on extracting road network graphs from satellite imagery.

\begin{figure}
	\begin{center}
		\includegraphics[width=1.0\linewidth]{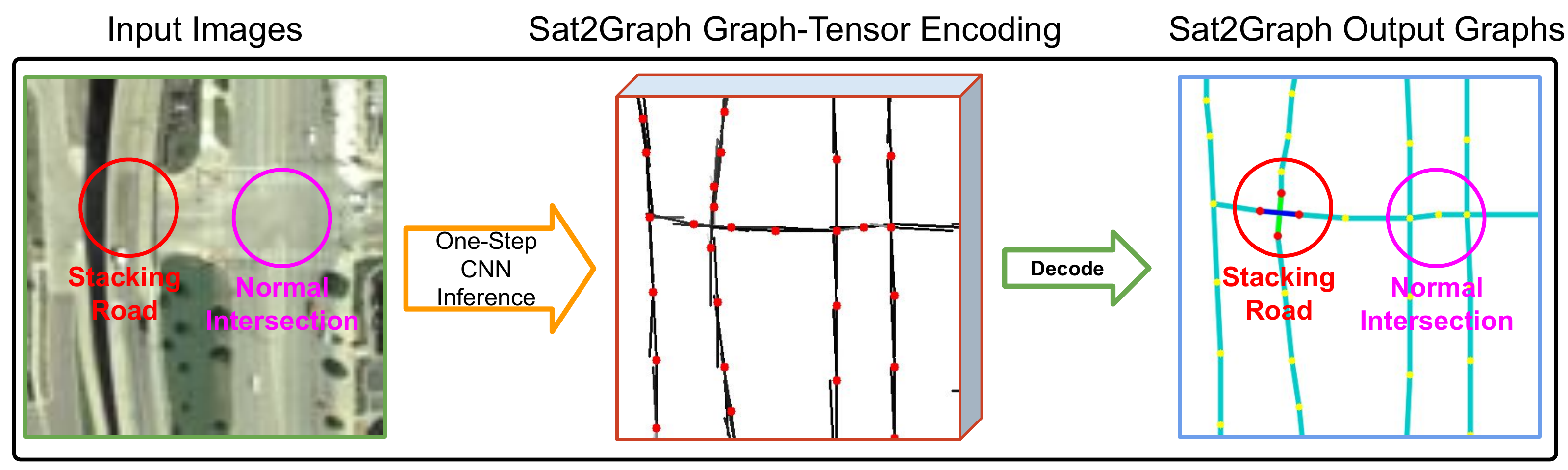}
	\end{center}
	\caption{Highlight of \name.}
	\label{fig:highlights}
\end{figure}

Although many techniques have been proposed~\cite{bastani2018roadtracer,batra2019improved,cheng2017automatic,chu2019neural,hinz2003automatic,li2018polymapper,mattyus2017deeproadmapper,wegner2015road,zhang2018road,zhou2018d,buslaev2018fully,Mosinska_2018_CVPR}, extracting road networks from satellite imagery is still a challenging computer vision task due to the complexity and diversity of the road networks.
Prior solutions fall into two categories: pixel-wise segmentation-based approaches and graph-based approaches. Segmentation-based approaches assign a \emph{roadness} score to each pixel in the satellite imagery. Then, they extract the road network graph using heuristic approaches. Here, the road segmentation acts as the intermediate representation of the road network graph. In contrast, graph-based approaches construct a road network graph directly from satellite imagery. Recently, Bastani \textit{et al.}~\cite{bastani2018roadtracer}, as well as several follow-up works~\cite{chu2019neural,li2018polymapper}, utilize graph-based solutions that iteratively add vertices and edges to the partially constructed graph.

We observe that the approaches in these two categories often tradeoff with each other. Segmentation-based approaches typically have a wider receptive field but rely on an intermediate non-graph representation and a post-processing heuristic (e.g., morphological thinning and line following) to extract road network graphs from this intermediate representation. The usage of the intermediate non-graph representation limits the segmentation-based approaches, and they often produce noisy and lower precision road networks compared with the graph-based methods as a result. To encourage the neural network model to focus more on the graph structure of road networks, recent work~\cite{batra2019improved} proposes to train the road segmentation model jointly with road directions, and the approach achieves better road connectivity through this joint training strategy.  However, a postprocessing heuristic is still needed.

In contrast, graph-based approaches~\cite{bastani2018roadtracer,chu2019neural,li2018polymapper} learn the graph structure directly. As a result, graph-based approaches  yield road network graphs with better road connectivity compared with the original segmentation-based approach~\cite{bastani2018roadtracer}.  However, the graph generation process is often iterative, resulting in a neural network model that focuses more on local information rather than global information. To take more global information into account, recent work~\cite{chu2019neural,li2018polymapper} proposes to improve the graph-based approaches with a sequential generative model, resulting in better performance compared with other state-of-art approaches. 

Recent advancements~\cite{batra2019improved,chu2019neural,li2018polymapper} in segmentation-based approaches and graph-based approaches respectively are primarily focused on overcoming the inherent limitations of their baseline approaches, which are exactly from the same aspects that the methods in the competing baseline approach (i.e., from the other category) claim as advantages. Based on this observation, a natural question to ask is if it is possible to combine the segmentation-based approach and the graph-based approach into one unified approach that can benefit from the advantages of both?

Our answer to this question is a new road network extraction approach, \name, which combines the inherent advantages of segmentation-based approaches and graph-based approaches into one simple, unified framework. To do this, we design a novel encoding scheme, {\em graph-tensor encoding} (GTE), to encode the road network graph into a tensor representation, making it possible to train a simple, non-recurrent, supervised model that predicts graph structures holistically from the input image. 

In addition to the tensor-based network encoding, this paper makes two contributions:
\begin{enumerate}
    \item \name\ surpasses state-of-the-art approaches in a widely used topology-similarity metric at all precision-recall trade-off positions in an evaluation over a large city-scale dataset covering 720 $km^2$ area in 20 U.S. cities and the popular SpaceNet roads dataset~\cite{van2018spacenet}. 
    \item \name\ can naturally infer stacked roads, which prior approaches don't handle.
\end{enumerate}

\section{Related work}
\textbf{Traditional Approaches.} Extracting road networks from satellite imagery has long history~\cite{fortier1999survey,wang2016review}. Traditional approaches generally use heuristics and probabilistic models to infer road networks from imagery. For examples, Hinz \textit{et al.}~\cite{hinz2003automatic} propose an approach to create road networks through a complicated road model that is built using detailed knowledge about roads and the environmental context, such as the nearby buildings, vehicles and so on. Wegner \textit{et al.}~\cite{wegner2015road} propose to model the road network with higher-order conditional random fields (CRFs). They first segment the aerial images into super-pixels, then they connect these super-pixels based on the CRF model. \\

\noindent\textbf{Segmentation-Based Approaches.} With the increasing popularity of deep learning, researchers have used convolutional neural networks (CNN) to extract road network from satellite imagery~\cite{zhang2018road, zhou2018d, buslaev2018fully, cheng2017automatic, mattyus2017deeproadmapper,batra2019improved}. For example, Cheng \textit{et al.}~\cite{cheng2017automatic} use an end-to-end cascaded CNN to extract road segmentation from satellite imagery. They apply a binary threshold to the road segmentation and use morphological thinning to extract the road center-lines. Then, a road network graph is produced through tracing the single-pixel-width road center-lines. Many other segmentation-based approaches proposed different improvements upon this basic graph extraction pipeline, including improved CNN backbones~\cite{buslaev2018fully,zhou2018d}, improved post-processing strategy~\cite{mattyus2017deeproadmapper}, improved loss functions~\cite{mattyus2017deeproadmapper,Mosinska_2018_CVPR}, incorporating GAN~\cite{zhang2019aerial,shi2017road,costea2017creating}, and joint training~\cite{batra2019improved}.

In contrast with existing segmentation-based approaches, \name\ does not rely on the road segmentation as intermediate representation and learns the graph structure directly. \\

\noindent\textbf{Graph-Based Approaches.} Graph-based approaches construct a road network graph directly from satellite imagery. Recently, Bastani \textit{et al.}~\cite{bastani2018roadtracer} proposed RoadTracer, a graph-based approach to generate road network in an iterative way. The algorithm starts from a known location on the road map. Then, at each iteration, the algorithm uses a deep neural network to predict the next location to visit along the road through looking at the surrounding satellite imagery of the current location. Recent works~\cite{chu2019neural,li2018polymapper} advanced the graph-based approach through applying sequential generative models (RNN) to generate road network iteratively. The usage of sequential models allows the graph generation model to take more context information into account compared with RoadTracer~\cite{bastani2018roadtracer}.  

In contrast with existing graph-based approaches, \name\ generates the road graphs in one shot (holistic). This allows \name\ to easily capture the global information and make better coordination of vertex placement. The non-recurrent property of \name\ also makes it easy to train and easy to extend (e.g., combine \name\ with GAN). We think this simplicity of \name\ is another advantage over other solutions. \\

\noindent\textbf{Using Other Data Sources and Other Digital Map Inference Tasks.}
Extracting road networks from other data sources has also been extensively studied, e.g., using GPS trajectories collected from moving vehicles~\cite{gpsbiagioni,gpsahmed2015comparison,gpsedelkamp2003route,gpsdavies2006scalable,gpscao2009gps,gpskharita,gpsroadrunner}. Besides road topology inference, satellite imagery also enables inference of different map attributes, including high-definition road details~\cite{mattyus2015enhancing, mattyus2016hd,he2019roadtagger}, road safety~\cite{najjar2017combining} and road quality~\cite{cadamuro2018assigning}.

\section{\name} 
In this section, we present the details of our proposed approach - \name. \name\ relies on a novel encoding scheme that can encode the road network graph into a three-dimensional tensor. We call this encoding scheme  Graph-Tensor Encoding (GTE). This graph-tensor encoding scheme allows us to train a simple, non-recurrent, neural network model to directly map the input satellite imagery into the road network graph (i.e., edges and vertices). As noted in the introduction, this graph construction strategy combines the advantages of segmentation-based and graph-based approaches. 

\begin{figure*}
	\begin{center}
		\includegraphics[width=1.0\linewidth]{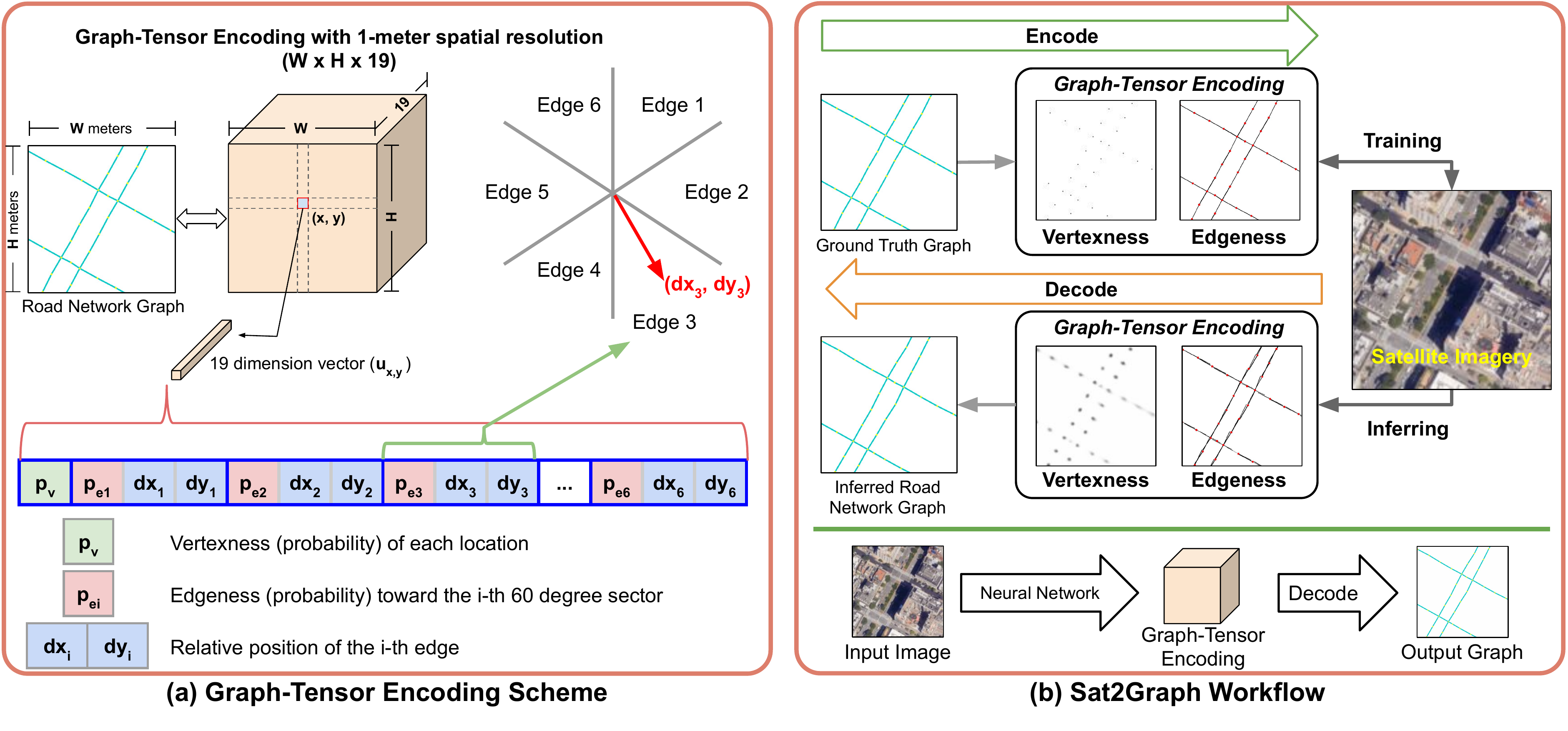}
	\end{center}
	\caption{Graph-Tensor Encoding and \name\ workflow.}
	\label{fig:overview}
\end{figure*}

\subsection{Graph-Tensor Encoding (GTE)}
\label{sec:graph-tensor-encoding}
We show our graph-tensor encoding (GTE) scheme in Figure \ref{fig:overview}(a). For a road network graph $G=\{V,E\}$ that covers a $W$ meters by $H$ meters region, GTE uses a $\frac{W}{\lambda}\times \frac{H}{\lambda} \times (1 + 3\cdot D_{max})$ 3D-tensor (denoted as $T$) to store the encoding of the graph. Here, the $\lambda$ is the spatial resolution, i.e., one meter, which restricts the encoded graph in a way that no two vertices can be co-located within a $\lambda \times \lambda$ grid, and $D_{max}$ is the maximum edges that can be encoded at each $\lambda \times \lambda$ grid.

The first two dimensions of $T$ correspond to the two spatial axes in the 2D plane. We use the vector at each spatial location $u_{x,y} = [T_{x,y,1},T_{x,y,2},...$ $,T_{x,y,(1 + 3\cdot D_{max})}]^T $ to encode the graph information. As shown in Figure \ref{fig:overview}(a), the vector $u_{x,y}$ has $(1 + 3\cdot D_{max})$ elements. Its first element $p_v \in [0,1]$ (vertexness) encodes the probability of having a vertex at position $(x,y)$. Following the first element are $D_{max}$ 3-element groups, each of which encodes the information of a potential outgoing edge from position $(x,y)$. For the $i$-th 3-element group, its first element $p_{e_i} \in [0,1]$ (edgeness) encodes the probability of having an outgoing edge toward $(dx_i, dy_i)$, i.e.,  an edge pointing from $(x,y)$ to $(x+dx_i, y+dy_i)$. Here, we set $D_{max}$ to six\dontshow{\san{six}} as we find that vertices with degree greater than six are very rare in road network graphs.

To reduce the number of possible different isomorphic encodings of the same input graph, GTE only uses the $i$-th 3-element group to encode edges pointing toward a $\frac{360}{D_{max}}$-degree sector from $(i-1)\cdot\frac{360}{D_{max}}$ degrees to $i\cdot\frac{360}{D_{max}}$ degrees. We show this restriction and an example edge (in red color) in Figure \ref{fig:overview}(a). This strategy imposes a new restriction on the encoded graphs -- for each vertex in the encoded graph, there can only be at most one outgoing edge toward each $\frac{360}{D_{max}}$-degree sector. However, we find this restriction does not impact the representation ability of GTE for most road graphs. This is because the graphs encoded by GTE are \textit{undirected}. We defer the discussion on this in Section~\ref{sec:discussion}.

\subsubsection{Encode}
Encoding a road network graph into GTE is straightforward. For road network extraction application, the encoding algorithm first interpolates the segment of straight road in the road network graph. It selects the minimum number of evenly spaced intermediate points so that the distance between consecutive points is under $d$ meters. This interpolation strategy regulates the length of the edge vector in GTE, making the training process stable. Here, a small $d$ value, e.g., $d<5$, converts GTE back to the road segmentation, making GTE unable to represent stacking roads. A very large $d$ value, e.g. $d=50$, makes the GTE hard to approximate curvy roads. For these reasons, we think a $d$ value between 15 to 25 can work the best. In our setup, we set $d$ to 20.

For stacked roads, the interpolation may produce vertices belonging to two overlapped road segments at the same position. When this happens, we use an iterative conflict-resolution algorithm to shift the positions of the endpoint vertices of the two edges. The goal is to make sure the distance between any two vertices (from the two overlapping edges) is greater than 5 meters. During training, this conflict-resolution pre-processing also yields more consistent supervision signal for stacked roads - overlapped edges tend to always cross near the middle of each edge. After this step, the encoding algorithm maps each of the vertices to the 3D-tensor $T$ following the scheme shown in Figure \ref{fig:overview}(a). For example, the algorithm sets the vertexness ($p_v$) of $u_{x,y}$ to 1 when there is a vertex at position $(x,y)$, otherwise the vertexness is set to 0.    

\subsubsection{Decode}
GTE's Decoding algorithm converts the predicted GTE (often noisy) of a graph back to the regular graph format (G = \{V,E\}). The decoding algorithm consists of two steps, (1) vertex extraction and, (2) edge connection. As both the vertexness predictions and edgeness predictions are real numbers between 0 and 1, we only consider vertices and edges with probability greater than a threshold (denoted as $p_{thr}$). 

In the vertex extraction step, the decoding algorithm extracts the potential vertices through localizing the local maximas of the vertexness map (we show an example of this in Figure~\ref{fig:overview}(b)). The algorithm only considers the local maximas with vertexness greater than $p_{thr}$. 

In the edge connection step, for each candidate vertex $v \in V$, the decoding algorithm connects its outgoing edges to other vertices.
For the $i$-th edge of vertex $v \in V$, the algorithm computes its distance to all nearby vertices $u$ through the following distance function, 
\begin{equation}
\begin{split}
    d(v,i, u) =  &||(v_x + dx_i, v_y + dy_i) - (u_x, u_y)||  \\
     &+ w\cdot cos_{dist} ( (dx_i, dy_i), (u_x - v_x, u_y - v_y))
\end{split}
\label{eq:dist}
\end{equation}
, where $cos_{dist}(v_1, v_2)$ is the cosine distance of the two vectors, and $w$ is the weight of the cosine distance in the distance function. Here, we set $w$ to a large number, i.e., 100, to avoid incorrect connections. After computing this distance, the decoding algorithm picks up a vertex $u^\prime$ that minimizes the distance function $d(v, i, u)$, and adds an edge between $v$ and $u^\prime$. We set a maximum distance threshold, i.e., 15 meters, to avoid incorrect edges being added to the graph when there are no good candidate vertices nearby.  

\subsection{Training \name}
We use cross-entropy loss (denoted as $\mathcal{L}_{CE}$) and $L_2$-loss to train \name. The cross-entropy loss is applied to vertexness channel ($p_v$) and edgeness channels ($p_{e_i}\  i\in\{1,2,...,D_{max}\}$), and the $L_2$-loss is applied to the edge vector channels ($(dx_i, dy_i)\  i\in\{1,2,...,D_{max}\}$). GTE is inconsistent along long road segments. In this case, the same road structure can be mapped to different ground truth labels in GTE representation. Because of this inconsistency, we only compute the losses for edgeness and edge vectors at position $(x,y)$ when there is a vertex at position $(x,y)$ in the ground truth. We show the overall loss function below ($\hat{T},\hat{p_v},\hat{p_{e_i}}, \hat{d_{x_i}},\hat{d_{y_i}}$are from ground truth),
\begin{equation}
\begin{split}
\mathcal{L}(T,\hat{T})& = \sum_{(x,y)\in[1..W]\times[1..H]} \Bigg( \mathcal{L}_{CE}(p_v, \hat{p_v}) \\
& + \hat{T}_{x,y,1} \cdot \Big(\sum_{i=1}^{D_{max}} \big( \mathcal{L}_{CE}(p_{e_i}, \hat{p_{e_i}}) 
+ \mathcal{L}_2((dx_i,dy_i),(\hat{dx_i},\hat{dy_i}))\ \big) \Big) \Bigg)\\
\end{split}
\end{equation}

In Figure~\ref{fig:overview}(b), we show the training and inferring workflows of \name.
\name\ is agnostic to the CNN backbones. In this paper, we choose to use the Deep Layer Aggregation (DLA)~\cite{yu2018deep} segmentation architecture as our CNN backbone. We use residual blocks~\cite{he2016deep} for the aggregation function in DLA. The feasibility of training \name\ with supervised learning is counter-intuitive because of the GTE's inconsistency. We defer the discussion of this to Section~\ref{sec:discussion}.

\section{Evaluation}
We now present experimental results comparing \name\ to several state-of-the-art road-network generation systems.

\subsection{Datasets}
We conduct our evaluation on two datasets, one is a large city-scale dataset and the other is the popular SpaceNet roads dataset~\cite{van2018spacenet}.

\textbf{City-Scale Dataset.} Our city-scale dataset covers 720 $km^2$ area in 20 U.S. cities. We collect road network data from OpenStreetMap~\cite{haklay2008openstreetmap} as ground truth and the corresponding satellite imagery through Google static map API~\cite{googleapi}. The spatial resolution of the satellite imagery is set to one meter per pixel. This dataset enables us to evaluate the performance of different approaches at city scale, e.g., evaluating the quality of the shortest path crossing the entire downtown of a city on the inferred road graphs. 

The dataset is organized as 180 tiles; each tile is a 2 km by 2 km square region.  We randomly choose 15\% (27 tiles) of them as a testing dataset and 5\% (9 tiles) of them as a validation dataset. The remaining 80\% (144 tiles) are used as training dataset.

\textbf{SpaceNet Roads Dataset.} Another dataset we used is the SpaceNet roads Dataset~\cite{van2018spacenet}. Because the ground truth of the testing data in the SpaceNet dataset is not public, we randomly split the 2549 tiles (non-empty) of the original training dataset into training(80\%), testing(15\%) and validating(5\%) datasets. Each tile is a 0.4 km by 0.4 km square. Similar to the city-scale dataset, we resize the spatial resolution of the satellite imagery to one meter per pixel.  

\subsection{Baselines}
We compare \name\ with four different segmentation-based approaches and one graph-based approach.\\

\noindent\textbf{Segmentation-Based Approaches.} We use four different segmentation-based approaches as baselines.
\begin{enumerate}
    \item \textit{Seg-UNet:} Seg-UNet uses a simple U-Net~\cite{ronneberger2015u} backbone to produce road segmentation from satellite imagery. The model is trained with cross-entropy loss. This scheme acts as the naive baseline as it is the most straightforward solution for road extraction.
    \item \textit{Seg-DRM~\cite{mattyus2017deeproadmapper}(ICCV-17):} Seg-DRM uses a stronger CNN backbone which contains 55 ResNet~\cite{he2016deep} layers to improve the road extraction performance. Meanwhile, Seg-DRM proposes to train the road segmentation model with soft-IoU loss to achieve better performance. However, we find training the Seg-DRM model with cross-entropy loss yields much better performance in terms of topology correctness. Thus, in our evaluation, we train the Seg-DRM model with cross-entropy loss.
    \item \textit{Seg-Orientation~\cite{batra2019improved}(ICCV-19):} Seg-Orientation is a recent state-or-the-art approach which proposes to improve the road connectivity by joint learning of road orientation and road segmentation. Similar to Seg-DRM, we show the results of Seg-Orientation trained with cross-entropy loss as we find it performs better compared with soft-IoU loss.   
    \item \textit{Seg-DLA:} Seg-DLA is our enhanced segmentation-based approach which uses the same CNN backbone as our \name\ model. Seg-DLA, together with Seg-UNet, act as the baselines of an ablation study of \name.   
\end{enumerate}

\noindent\textbf{Graph-Based Approaches.} For graph-based approaches, we compare our \name\ solution with RoadTracer~\cite{bastani2018roadtracer}(CVPR-18) by applying their code on our dataset. During inference, we use peaks in the segmentation output as starting locations for RoadTracer's iterative search.

\subsection{Implementation Details}

\textbf{Data Augmentation:} For all models in our evaluation, we augment the training dataset with random image brightness, hue and color temperature, random rotation of the tiles, and random masks on the satellite imagery.  \\ 

\noindent\textbf{Training:} We implemented both \name\ and baseline segmentation approaches using Tensorflow. We train the model on a V100 GPU for 300k iterations (about 120 epochs) with a learning rate starting from 0.001 and decreasing by 2x every 50k iterations. We train all models with the same receptive field, i.e., 352 by 352. 
We evaluate the performance on the validation dataset for each model every 5k iterations during training, and pick up the best model on the validation dataset as the converged model for each approach to avoid overfitting.\\ 

\subsection{Evaluation Metrics}
In the evaluation, we focus on the topology correctness of the inferred road graph rather than edge-wise correctness. This is because the topology correctness is often crucial in many real-world applications. For example, in navigation applications, a small missing road edge in the road graph could make two regions disconnected. This small missing road segment is a small error in terms of edge-wise correctness but a huge error in terms of topology correctness.  

We evaluate the topology correctness of the inferred road graphs through two metrics,  TOPO~\cite{biagioni2012inferring} and APLS~\cite{van2018spacenet}. Here, we describe the high level idea of these two metrics. Please refer to~\cite{biagioni2012inferring,van2018spacenet} for more details about these two metrics.  \\

\noindent\textbf{TOPO metric:} TOPO metric measures the similarity of sub-graphs sampled on the ground truth graph and the inferred graph from a seed location. The seed location is matched to the closest seed node on each graph. Here, given a seed node on a graph, the sub-graph contains all the nodes such that their distances (on the graph) to the seed node are less than a threshold, e.g., 300 meters. For each seed location, the similarity between two sampled sub-graphs is quantified as precision, recall and $F_1$-score. The metric reports the average precision, recall and $F_1$-score over randomly sampled seed locations over the entire region.         

The TOPO metric has different implementations. We implement the TOPO metric in a very strict 
way following the description in \cite{gpsroadrunner}. This strict implementation allows the metric to penalize detailed topology errors. 

\noindent\textbf{APLS metric:} APLS measures the quality of the shortest paths between two locations on the graph. For example, suppose the shortest path between two locations on the ground truth map is 200 meters, but the shortest path between the same two locations on the inferred map is 20 meters (a wrong shortcut), or 500 meters, or doesn't exist. In these cases, the APLS metric yields a very low score, even though there might be only one incorrect edge on the inferred graph.

\subsection{Quantitative Evaluation}

\begin{figure}
	\begin{center}
		\includegraphics[width=1.0\linewidth]{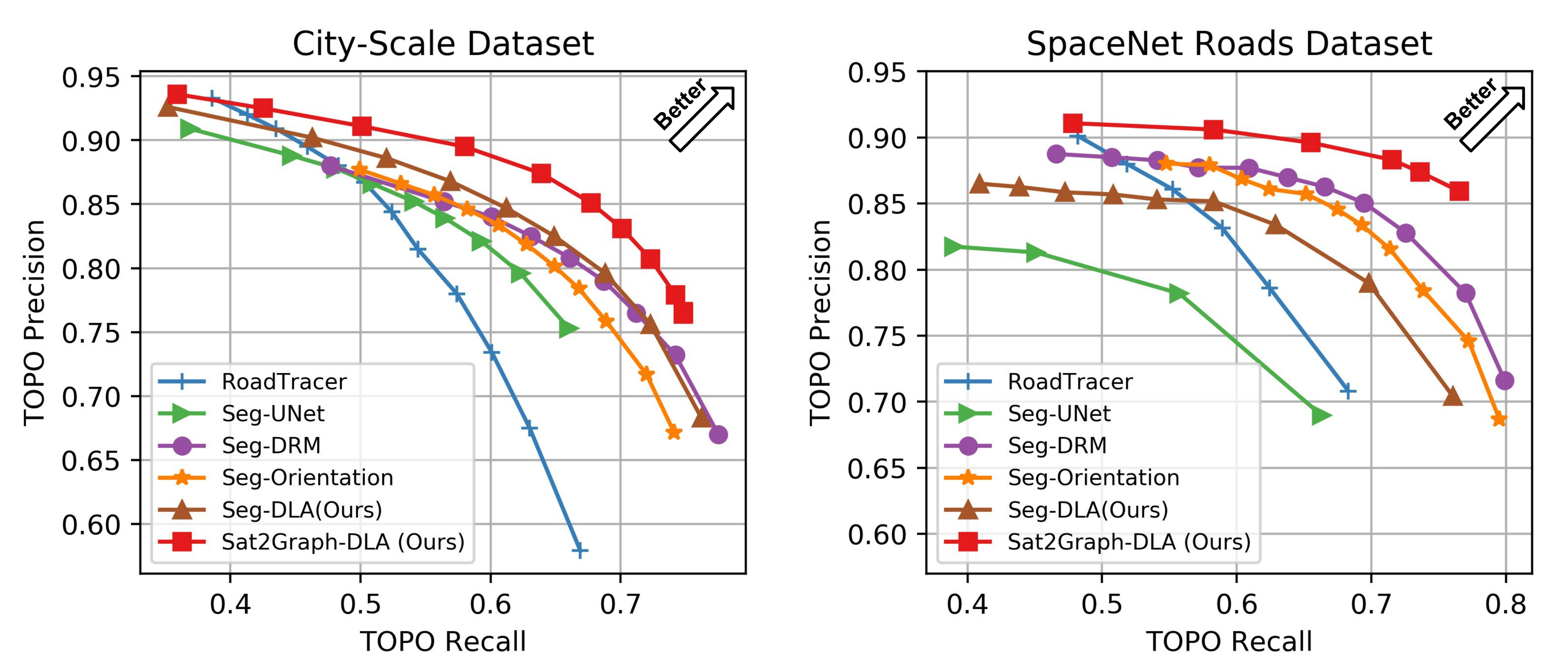}
	\end{center}
	\caption{TOPO metric precision-recall trade-off curves}
	\label{fig:topo}
\end{figure}

\textbf{Overall Quality.} Each of the approaches we evaluated has one major hyper-parameter, which is often a probability threshold, that allows us to make different precision-recall trade-offs. We change this parameter for each approach to plot an precision-recall curve. We show the precision-recall curves for different approaches in Figure \ref{fig:topo}. This precision-recall curve allows us to see the full picture of the capability of each approach. We also show the best achievable TOPO $F_1$-score and APLS score of each approach in Table \ref{table:topo} for reference.

From Figure~\ref{fig:topo}, we find an approach may not always better than another approach at different precision-recall position (TOPO metric). For examples, the graph-based approach RoadTracer performs better than others when the precision is high, whereas the segmentation-based approach DeepRoadMapper performs better when the recall is high. 

Meanwhile, we find an approach may not always better than another approach on both TOPO and APLS. For example, in Table~\ref{table:topo}, RoadTracer has the best APLS score but the worst TOPO $F_1$-score in the five baselines on the city-scale dataset. This is because RoadTracer is good at coarse-grained road connectivity and the precision of inferred road graphs rather than recall. For example, in Figure~\ref{fig:largetiles}(a), RoadTracer is better compared with Seg-DRM and Seg-Orientation in terms of road connectivity when the satellite imagery is full of shadow.  

In contrast, \name\ surpasses all other approaches on APLS metric and at all TOPO precision-recall positions --  for a given precision, \name\ always has the best recall; and for a given recall, \name\ always has the best precision. We think this is because \name 's graph-tensor encoding takes advantages from both the segmentation-based approaches and graph-based approaches, and allows \name\ to infer stacking roads that none of the other approaches can handle. As an ablation study, we compare \name-DLA with Seg-DLA (Seg-DLA uses the same CNN backbone as \name-DLA). We find the superiority of \name\ comes from the graph-tensor encoding rather than the stronger CNN backbone.   

\begin{table}
	\centering
	\begin{tabular}{|l||c|c|c|c||c|c|c|c|} 
		\hline
		Method &\multicolumn{4}{c||}{City-Scale Dataset} & \multicolumn{4}{c|}{SpaceNet Roads Dataset} \\ \cline{2-9}
		&\ Prec.\ &\ Rec.\ & $F_1$ & APLS &\ Prec.\ &\ Rec.\ & $F_1$ & APLS  \\ \hline 
	    RoadTracer\cite{bastani2018roadtracer}(CVPR-18) & 78.00 & 57.44 & 66.16 & 57.29 &78.61&62.45&69.60&56.03  \\ 
	    Seg-UNet & 75.34 & 65.99 & 70.36 & 52.50 & 68.96 & 66.32 & 67.61 & 53.77  \\ 
	    Seg-DRM\cite{mattyus2017deeproadmapper}(ICCV-17) & 76.54 & 71.25 & 73.80 & 54.32  &82.79 & 72.56 & 77.34 & 62.26  \\ 
	    Seg-Orientation\cite{batra2019improved}(ICCV-19) &75.83&68.90&72.20 & 55.34 &81.56&71.38 &76.13 &58.82 \\ \hline 
	    Seg-DLA(ours) & 75.59 & 72.26 & 73.89 & 57.22 &78.99 &69.80 &74.11 & 56.36 \\ 
	    Sat2Graph-DLA(ours) &  80.70 & 72.28 &\textbf{76.26} & \textbf{63.14} &85.93 &76.55 &\textbf{80.97} & \textbf{64.43} \\ \hline
	\end{tabular}
	\caption{Comparison of the \textit{best achievable} TOPO $F_1$-score and APLS score. We show the best TOPO $F_1$-score's corresponding precision and recall just for reference not for comparison. (All the values in this table are percentages) }
	\label{table:topo}
\end{table}

\textbf{Benefit from GTE.} In addition to the results shown in Table~\ref{table:topo}, we show the results of using GTE with other backbones. On our city-wide dataset, we find GTE can improve the TOPO $F_1$-score from 70.36\% to 76.40\% with the U-Net backbone and from 73.80\% to 74.66\% with the Seg-DRM backbone. Here, the improvement on Seg-DRM backbone is minor because Seg-DRM backbone has a very shallow decoder.

\textbf{Sensitivity on $w$.} In our decoding algorithm, we have a hyper-parameter $w$ which is the weight of the cosine distance term in equation~\ref{eq:dist}. In Table~\ref{table:w}, we show how this parameter impacts the TOPO $F_1$-score on our city-wide dataset. We find the performance is robust to w - the $F_1$-scores are all greater than 76.2\% with w in the range from 5 to 100.
\begin{table}
\centering
\begin{tabular}{|l|c|c|c|c|c|c|c|} 
\hline
Value of $w$ & 1 & 5 & 10 & 25 & 75 & 100 & 150 \\ \hline 
$F_1$-score & 75.87\% & 76.28\% & 76.62\% & 76.72\% & 76.55\% & 76.26\% & 75.68\% \\ \hline
\end{tabular}
\caption{TOPO $F_1$ scores on our city-wide dataset with different $w$ values.}
\label{table:w}
\end{table}

\textbf{Vertex threshold and edge threshold.} In our basic setup, we set the vertex threshold and the edge threshold of \name\ to the same value. However, we can also use independent probability thresholds for vertices and edges. We evaluate this by choosing a fixed point and vary one probability threshold at a time. We find the vertex threshold dominates the performance and using a higher edge probability threshold (compared with the vertex probability) is helpful to achieve better performance. 

\textbf{Stacking Road.} We evaluate the quality of the stacking road by matching the overpass/underpass crossing points between the ground truth graphs and the proposed graphs. In this evaluation, we find our approach has a precision of 83.11\% (number of correct crossing points over the number of all proposed crossing points) and a recall of 49.81\% (number of correct crossing points over the number of all ground-truth crossing points) on stacked roads. In fact only 0.37\% of intersections are incorrectly predicted as overpasses/underpasses (false-positive rate). We find some small roads under wide highway roads are missing entirely. We think this is the reason for the low recall. 

\subsection{Qualitative Evaluation}
\begin{figure}[h]
	\begin{center}
		\includegraphics[width=1.0\linewidth]{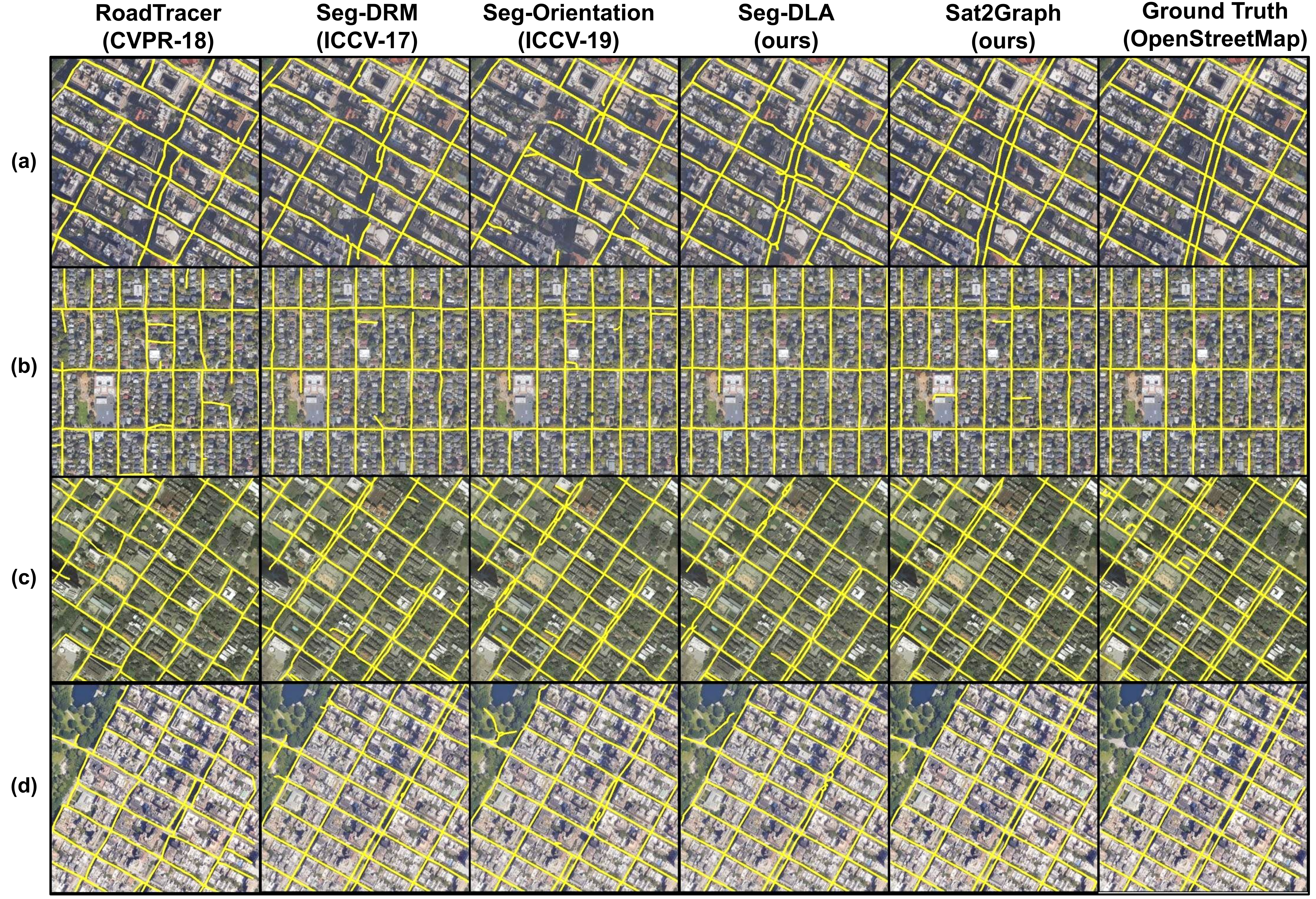}
	\end{center}
	\caption{Qualitative Comparison in Regular Urban Areas. We use the models that yield the best TOPO $F_1$ scores to create this visualization.\dontshow{ \srm{cyan lines against green and blue backgrounds are hard to see -- can you boost the contrast somehow, e.g., by desaturating / lightening underlying maps, or using thicker lines?}\songtao{Increased the line width a little bit and decreased the contrast of the background satellite imagery, and yellow seems to be more printer-friendly}}}
	\label{fig:largetiles}
\end{figure}

\textbf{Regular Urban Areas.} In the regular urban areas (Figure~\ref{fig:largetiles}), we find the existing segmentation-based approach with a strong CNN backbone (Seg-DLA) and better data augmentation techniques has already been able to achieve decent results in terms of both precision and recall, even if the satellite imagery is full of shadows and occlusions. Compared with \name, the most apparent remaining issue of the segmentation-based approach appears at parallel roads. We think the root cause of this issue is from the fundamental limitation of segmentation-based approaches --- the road-segmentation intermediate representation. \name\ eliminates this limitation through graph-tensor encoding, thereby, \name\ is able to produce detailed road structures precisely even along closeby parallel roads.\\

\begin{figure}[t]
	\begin{center}
		\includegraphics[width=1.0\linewidth]{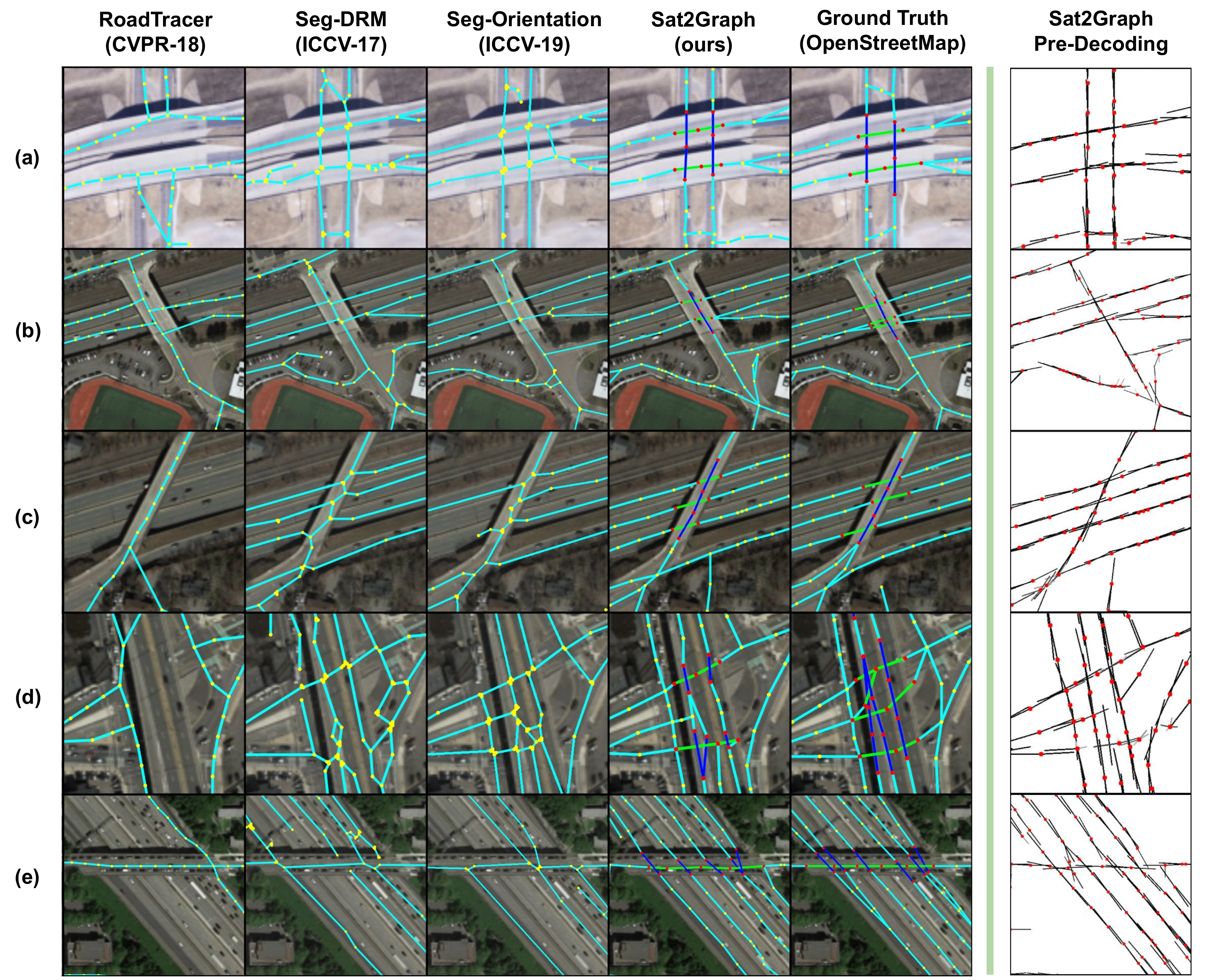}
	\end{center}
	\caption{Qualitative Comparison for Stacked Roads. \name\ robustly infers stacked roads in examples (a-c), but makes some errors in (d) and (e). Prior work infers only planar graphs and incorrectly captures road topology around stacked roads in all cases. We highlight edges that cross without connecting in green and blue.
	}
	\label{fig:stackingroads}
\end{figure}

\noindent\textbf{Stacked Roads.} We show the orthogonal superiority of \name\ on stacked roads in Figure~\ref{fig:stackingroads}. None of the existing approaches can handle stacked roads, whereas \name\ can naturally infer stacked roads thanks to the graph-tensor encoding. We find \name\ may still fail to infer stacking roads in some complicated scenarios such as in Figure~\ref{fig:stackingroads}(d-e). We think this can be further improved in a future work, such as adding discriminative loss to regulate the inferred road structure.

\section{Discussion}
\label{sec:discussion}
There are two concerns regarding \name:
(1) it seems that the heavily restricted and non-lossless graph-tensor encoding may not be able to correctly represent all different road network graphs, and (2) training a model to output GTE representation with supervised learning seems impossible because the GTE representation is not consistent.

\textbf{Concern about the encoding capability.} We think there are two reasons that make GTE able to encode almost all road network graphs.

First, the road network graph is \textit{undirected.} Although roads have directions, the road directions can be added later as road attributes, after the road network extraction. In this case, for each edge $e=(v_a,v_b)$, we only need to encode one link from $v_a$ to $v_b$ or from $v_b$ to $v_a$, rather than encode both of the two links. Even though GTE has the $\frac{360}{D_{max}}$-degree sector restriction on outgoing edges from one vertex, this undirected-graph property makes it possible to encode very sharp branches such as the branch vertices between a highway and an exit ramp.

Second, the road network graph is \textit{interpolatable.} There could be a case where none of the two links of an edge $e=(v_a,v_b)$ can be encoded into GTE because both $v_a$ and $v_b$ need to encode their other outgoing links. However, because the road network graph is interpolatable, we can always interpolate the edge $e$ into two edges $e_1 = (v_a, v^\prime)$ and $e_2 = (v^\prime, v_b)$. After the interpolation, the original geometry and topology remain the same but we can use the additional vertex $v^\prime$ to encode the connectivity between $v_a$ and $v_b$.

In Table~\ref{table:dmax}, we show the ratios of edges that need to be fixed using the \textit{undirected} and \textit{interpolatable} properties in our dataset with different $D_{\text{max}}$ values.  

\begin{table}
\centering
\begin{tabular}{|l|c|c|c|c|c|} 
\hline
$D_{\text{max}}$& 3 & 4 & 5 & 6 & 8 \\ \hline 
Fixed with the \textit{undirected} property& 8.62\% & 2.81\% & 1.18\% & 0.92\% & 0.59\% \\ 
Fixed with the \textit{interpolatable} property& 0.013\% & 0.0025\% & 0.0015\% & 0.0013\% & 0.0013\% \\ \hline
\end{tabular}
\caption{}
\label{table:dmax}
\end{table}

\textbf{Concern about supervised learning.} Another concern with GTE is that for one input graph, there exist many different isomorphic encodings for it (e.g., there are many possible vertex interpolations on a long road segment.). These isomorphic encodings produce inconsistent ground truth labels. During training, this inconsistency of the ground truth can make it very hard to learn the right mapping through supervised learning.  

However, counter-intuitively, we find \name\ is able to learn through supervised learning and learn well. We find the key reason of this is because of the inconsistency of GTE representation doesn't equally impact the vertices and edges in a graph. For example, the locations of intersection vertices are always consistent in different isomorphic GTEs. 

We find GTE has high label consistency for supervised learning 
at important places such as intersections and overpass/underpass roads. Often, these places are the locations where the challenges really come from. Although GTE has low consistency for long road segments, the topology of the long road segment is very simple and can still be corrected through GTE's decoding algorithm.

\section{Conclusion}
In this work, we have proposed a simple, unified road network extraction solution that combines the advantages from both segmentation-based approaches and graph-based approaches.  Our key insight is a novel graph-tensor encoding scheme. Powered by this graph-tensor approach, \name\ is able to surpass existing solutions in terms of topology-similarity metric at all precision-recall points in an evaluation over two large datasets. 
Additionally, \name\ naturally infers stacked roads like highway overpasses that none of the existing approaches can handle.

\clearpage
%
%
\bibliographystyle{splncs04}
\bibliography{egbib}
\end{document}